\documentclass[10 pt, conference]{ieeeconf}  
\IEEEoverridecommandlockouts                              
\usepackage{graphics}    
\usepackage{times}       

\usepackage{caption}
\usepackage{subcaption}

\usepackage{amsmath,amssymb,amsfonts}
\usepackage{graphicx}
\usepackage{textcomp}
\usepackage{xcolor}

\usepackage{siunitx}

\usepackage[font=small]{caption}

\def\figref#1{Fig.~\ref{#1}}

\def\eqref#1{Eq.~(\ref{#1})}

\title{\LARGE \bf Context-Based Meta Reinforcement Learning 
\\for Robust and Adaptable Peg-in-Hole Assembly Tasks}

\author{ Ahmed Shokry$^{1,2}$ \hspace{15pt} \and Walid Gomaa$^{3,4}$ \hspace{15pt} \and Tobias Zaenker$^1$ \hspace{15pt} \and Murad Dawood $^{1,2}$   \and \hspace{15pt} Rohit Menon$^{1,2}$ \and \hspace{35pt} Shady A. Maged$^{5}$ \hspace{30pt} \and Mohammed I. Awad$^5$  \hspace{35pt}  \and  Maren Bennewitz$^{1,2}$    
  \thanks{$^1$Humanoid Robots Lab and Center for Robotics, University of Bonn, Germany. $^2$Lamarr Institute for Machine Learning and Artificial Intelligence, Bonn, Germany. $^3$Cyber Physical Systems Lab, Egypt Japan University of Science and Technology, Alexandria, Egypt. $^4$Faculty of Engineering, Alexandria University, Alexandria, Egypt. $^5$Mechatronics Department, Ain Shams University, Cairo, Egypt.}
  \thanks{
This work has been supported by the BMBF within the Robotics Institute Germany, grant No. 16ME0999.
  }%
}

\begin{document}
\maketitle
\thispagestyle{empty} 
\pagestyle{empty}


\newcommand{\wg}[1]{\textcolor{blue}{#1}}

\begin{abstract}
Autonomous assembly is an essential capability for industrial and service robots, with Peg-in-Hole (PiH) insertion being one of the core tasks. However, PiH assembly in unknown environments is still challenging due to uncertainty in task parameters, such as the hole position and orientation, resulting from sensor noise. 
Although context-based meta reinforcement learning~(RL) methods have been previously presented to adapt to unknown task parameters in PiH assembly tasks, the performance depends on a sample-inefficient procedure or human demonstrations. Thus, to enhance the applicability of meta RL in real-world PiH assembly tasks, we propose to train the agent to use information from the robot's forward kinematics and an uncalibrated camera. Furthermore, we improve the applicability by efficiently adapting the meta-trained agent to use data from force/torque sensor. Finally, we propose an adaptation procedure for out-of-distribution tasks whose parameters are different from the training tasks. Experiments on simulated and real robots prove that our modifications enhance the sample efficiency during meta training, 
real-world adaptation performance, 
and generalization of the context-based meta RL agent in PiH assembly tasks compared to previous approaches.

\end{abstract}

\section{Introduction}



\begin{figure}[t]

\centerline{\includegraphics[width=0.47\textwidth]{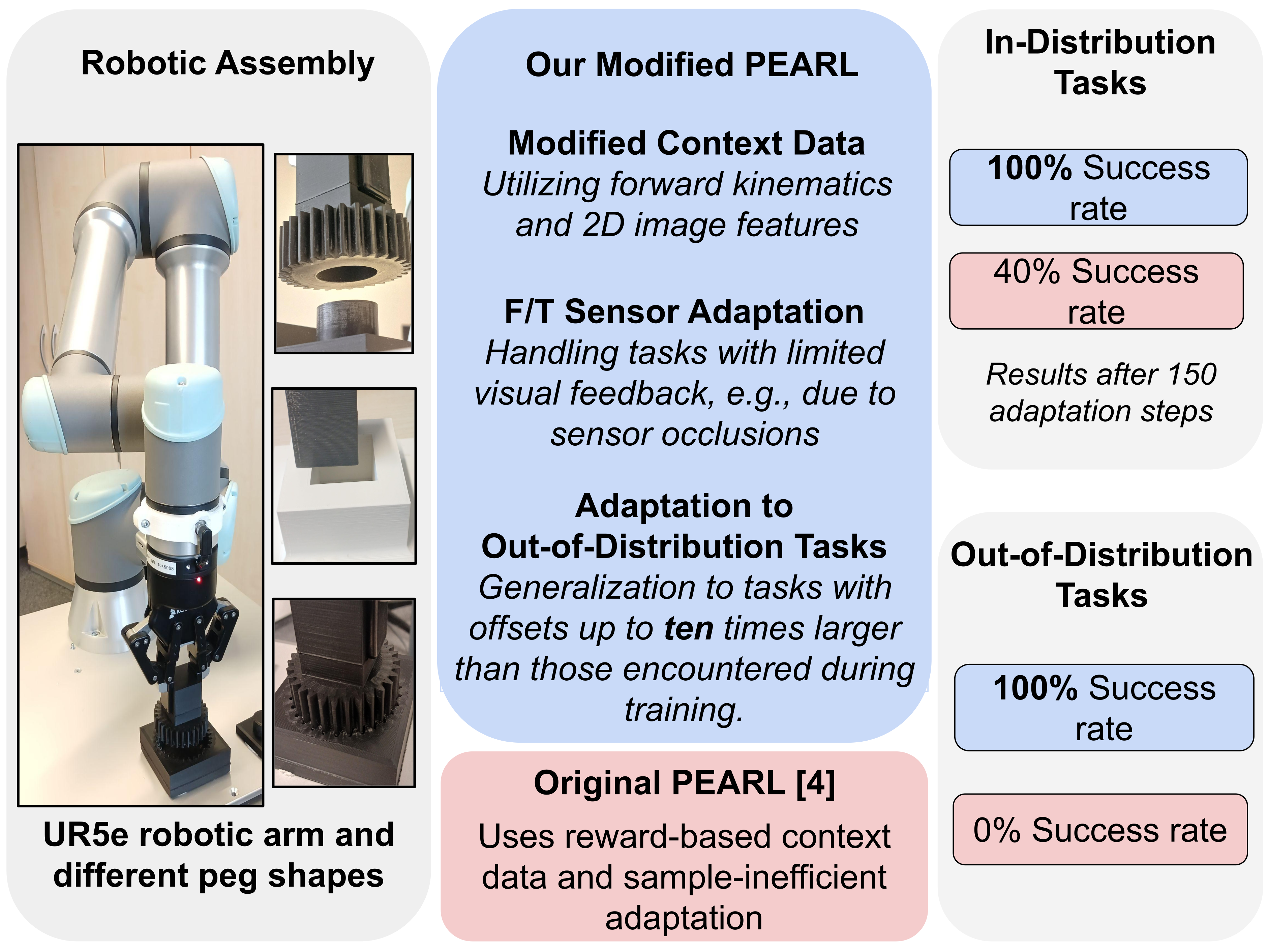}}
\caption{
We use context-based meta reinforcement learning  to perform peg-in-hole assembly tasks with unknown hole position. Unlike previous work~\cite{metarlinsertion}, which uses immeasurable reward as a part of the context data to infer the task, which results in a sample-inefficient adaptation, we use data from the robot's forward kinematics and uncalibrated camera to infer task parameters. Additionally, we adapt the agent to use force/torque sensor data to avoid occlusion problems.  Finally, we propose an adaptation procedure to out-of-distribution tasks with huge errors in the estimated hole position and demonstrate the superior performance of our methods.}
\label{fig:cover_fig}
\end{figure}

Assembly tasks are widely used in industry~\cite{review_of_robotic_assembly_2022}, service and collaborative robots \cite{collabrotive_robots}, and space technology \cite{space_assembly}. 
While robotic manipulation tasks such as grasping and pushing have seen increasing success in real-world applications, assembly tasks remain particularly challenging due to the high precision and adaptability required. 
Peg-in-hole (PiH) insertion~(see~\figref{fig:cover_fig}) exemplifies this challenge, as it requires real-time adaptation and robustness to uncertainties in position and contact dynamics.

PiH assembly has been studied in the literature as being the core of robotic assembly tasks \cite{ review_assembly_tasks_2019, compare_contact_model_based_and_model_free}. 
However, PiH assembly with uncertain parameters such as the hole position and orientation, resulting from sensor noise, remains challenging~\cite{review_of_robotic_assembly_2022}. 
Meta reinforcement learning (RL) offers a promising paradigm by enabling robots to rapidly adapt to new insertion tasks and varying conditions using prior experience. 
Specifically, context-based meta RL offers a robust solution by learning to infer the unknown parameters from task data and adapting the policy accordingly~\cite{metarl_survey,pearl}. 

In particular, context-based meta RL utilizes a context encoder to infer task-specific information and generate a latent representation that adapts the policy. Typically, task inference relies on rewards as part of the context data. However, in real-world scenarios, rewards may not always be available during deployment, making it difficult for the agent to accurately infer the task. As a result, the agent resorts to randomly sampling from the latent representation learned during meta-training, leading to inconsistent adaptation performance, and reduced sample efficiency, due to the many trials needed for sampling the correct representation \cite{pearl,metarl_survey}. 

Prior work on context-based meta RL for PiH assembly tasks \cite{metarlinsertion} has shown promising results but relies on reward signals that are unmeasurable in real-world settings, limiting their applicability and adaptation performance. A potential alternative is to replace reward-based context inference with human demonstrations \cite{offline_meta_rl_assembly}. However, human demonstrations may not always be readily available, especially in scenarios where autonomous adaptation is required. This underscores the need for alternative strategies that enable sample-efficient adaptation while ensuring consistent and reliable task execution in uncertain environments, see~Fig. \ref{fig:cover_fig}.

In this paper, (i) we propose the modification of the context data to use forward kinematics and easily measurable data from an uncalibrated vision sensor to enhance the real-world adaptation performance to address challenges arising from limited visual feedback, e.g., due to sensor occlusions, (ii) we develop a framework that enables the meta-trained agent to use alternative sensory modalities, such as force/torque feedback, (iii) we present an adaptation procedure to out-of-distribution (OOD) tasks, with parameters different from those of the training tasks, achieving generalization and solving one of the main limitations of context-based meta RL algorithms~\cite{metarl_survey}.

We present extensive simulated as well as real-world experiments with different peg and hole shapes that demonstrate the superior performance of our proposed modifications compared to prior work~\cite{metarlinsertion} regarding the training and adaptation efficiency, robustness to more uncertainty sources, real-world adaptation performance, and generalization.



\begin{figure}[t]
\centerline{\includegraphics[width=0.42\textwidth]{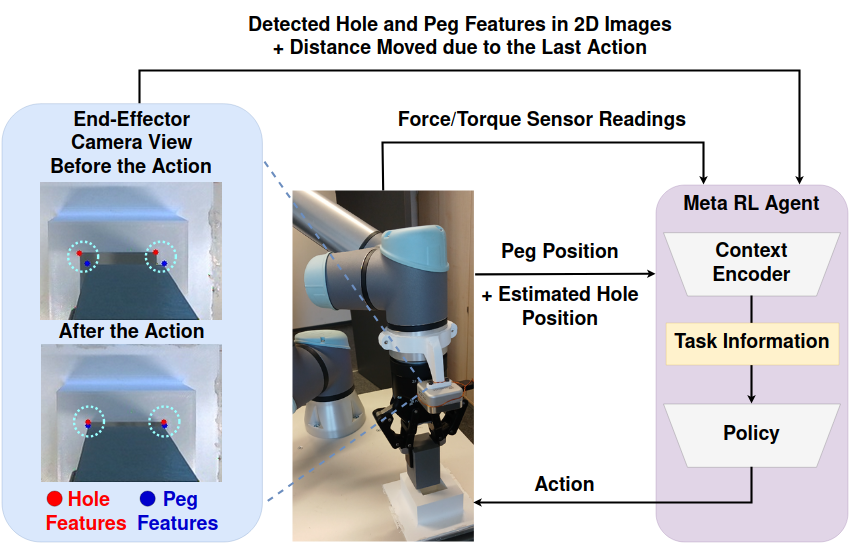}}
\caption{The meta RL agent receives the hole position estimated from noisy external sensor, the current peg position~(\textbf{middle}), the distance moved due to the last action calculated from forward kinematics, the detected hole and peg features in the 2D images captured before and after the action~(\textbf{left}), and the force/torque sensor reading from the robot. The context encoder~(\textbf{right}) uses these data to estimate the unknown information about the task, i.e., the actual hole position, and adapts the policy which produces the incremental robot motion. }
\label{fig:introductory_fig}
\end{figure}
\section{Related Work}
\label{sec:related_work}

Vision \cite{coarse_to_fine_visual_servoing,visual_spatial_attention_for_concrete_peg_in_hole_assembly} and force/torque sensors have been used to estimate the hole pose and perform the assembly \cite{review_of_robotic_assembly_2022}. Vision sensors require precise calibration and an additional heuristic-based search step \cite{spiralsearch} to compensate for residual errors \cite{fast_robust_peg_in_hole_insertion_with_continuous_visual_servoing,learning_based_visual_servoing,vision_based_6d_pih_for_practical_connectors_and_usb}. Although earlier works have achieved high accuracy in detecting hole features in 2D images, estimating the 3D hole pose suffers from high residual errors due to noisy depth sensing \cite{problems_depth_sensors_1,problems_depth_sensors_2}. Hence, we adapt the meta RL agent to solely rely on the hole features in the 2D image without requiring depth information or accurate camera calibration. 
 Unlike traditional visual servoing \cite{visual_servoing}, which requires approximate or estimated depth values and intrinsic parameters in addition to image features to control robot motion, our agent relies solely on image features. Furthermore, it leverages features detected over all previous time steps to control robot motion, which enhances robustness to instantaneous perturbations, such as occlusions occurring at certain times, compared to visual servoing that depends only on the current image features to decide the robot motion.

Force/torque sensors \cite{reducing_uncertainty_in_pose_estimation_using_force_data,force_torque_hole_pose_estimation_1,force_torque_hole_pose_estimation_2} have also been used to estimate the hole pose. However, they suffer from the sim-to-real gap~\mbox{\cite{factory,forge}}, which limits their usage in learning-based agents trained in simulation. Therefore, we propose a sample-efficient method that adapts the meta RL agent, trained in simulation, to effectively utilize force/torque sensor data using a limited amount of real-world samples.

RL has been proposed to learn robot actions directly from raw camera observations for assembly tasks~\cite{rl_vision_assembly_3,deep_visomotor_rl_assembly}, however, its generalization to changes in camera position and task configurations, such as peg shape and color, is limited. That is why, we use the detected features rather than raw pixels to train the agent to ensure its generalization to different task configurations.  Meta RL has been used to perform assembly tasks with uncertainty in the hole pose and different peg shapes. In~\cite{metarlinsertion}, the meta RL agent is trained in simulation on a distribution of tasks with different but known hole positions and consequently known rewards. The agent is then tested in the real world on tasks with an unknown hole position. Due to the absence of the dense reward in the real world, the agent uses a sample inefficient procedure to adapt to new tasks. In \cite{offline_meta_rl_assembly}, the meta RL agent uses human demonstrations to adapt to tasks with small uncertainty in the hole pose and different peg shapes. However, human demonstrations may not always be available. Therefore, we adapt the meta RL agent to use data that can be easily measured in the real world using onboard sensors.

Adapting the meta RL agent to use data from a different distribution, i.e., a different sensor, has also been proposed in \cite{rma}. A new context encoder is trained using supervised learning to produce the same output as the meta-trained encoder. However, the meta-trained encoder uses privileged information about the task and the new encoder uses a sequence of observations and actions as input and they produce a deterministic latent representation as output.


\section{Problem Description}
\label{sec:problem_definition}

This work considers PiH assembly tasks with unknown hole position (see Fig. \ref{fig:introductory_fig}). 
The goal is to train an agent that quickly infers the accurate hole position and controls the motion of a robotic manipulator to insert the peg grasped by the end-effector into the hole. 
We assume that an external sensor such as a camera provides the estimated hole position, but only with an error of $\pm\delta$ in the horizontal plane, similar to~\cite{metarlinsertion}.  Furthermore, we assume that the  manipulator is equipped with an eye-in-hand vision sensor and a force torque sensor at the wrist for context adaptation.

\section{Approach }
\label{sec: Approach}

To solve the Peg-in-Hole assembly task with unknown hole position, we utilize the context-based meta RL algorithm PEARL (Probabilistic Embeddings for Actor-critic RL)~\cite{pearl}, which is well-suited for this problem as it enables efficient task inference by leveraging context data collected during interactions. In our case, the agent has to infer the actual hole position from the context data and adjust its policy accordingly.
In this section, we first provide an overview of the PEARL algorithm and then introduce our proposed modifications to improve training and adaption efficiency, real-world applicability, and generalization to OOD tasks.

\subsection{PEARL Algorithm}
\label{sec:PEARL}

\begin{figure}[t]

\centerline{\includegraphics[scale=0.2]{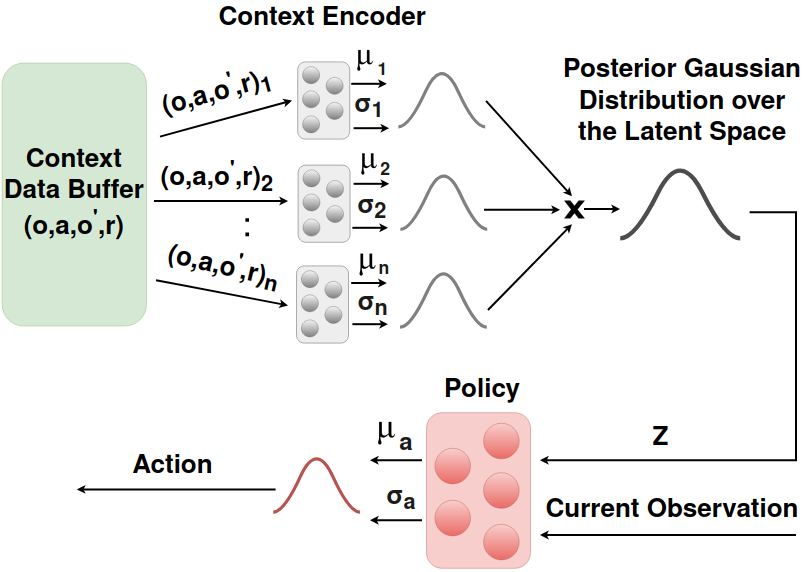}}
\caption{The context encoder neural network of PEARL~\cite{pearl} maps the collected context data $c=(o,a,o^\prime,r)$ to a posterior Gaussian distribution representing the agent's belief over the current task, from which latent variables are sampled to adapt the policy. }
\label{fig:pearl_algorithm}
\end{figure}

PEARL~\cite{pearl}  infers the current task using context data $c=(o,a,o^\prime,r)$, where $o$ is the observation, $o^\prime$ the next observation, $a$ the action, and $r$ the reward.
As shown in Fig. \ref{fig:pearl_algorithm}, the context data is fed to the context encoder to produce the mean and variance of a Gaussian distribution. After each collected trajectory, the produced distributions from all context data collected so far are multiplied together to produce one posterior Gaussian distribution over the latent space. The standard deviation represents the agent's uncertainty about the task. The latent variables sampled from this distribution are fed to the policy to adapt its behavior to the current task. The context encoder is trained to reconstruct the state action Q-value and the policy is trained to minimize the KL-divergence with the distribution over the Q-values. Since the context encoder reconstructs the Q-value that is used to train the policy, the context encoder is considered to implicitly reconstruct the policy's behavior. As a result, representations close to each other in the latent space produce similar actions and policy's behaviors and vice-versa.

\subsubsection{Observation, Reward, and Action Definitions}

\begin{figure}[t]

\centerline{\includegraphics[scale=0.17]{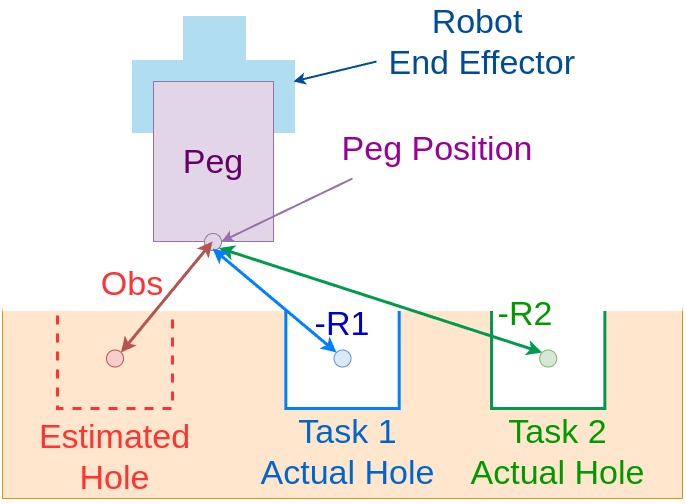}}
\caption{The observation is the peg position relative to the estimated hole. The reward is the negative $L_2$ distance between the peg and the actual hole position. Different actual hole positions have different rewards but the same observation \cite{metarlinsertion}. }
\label{pearl_obs_reward_definitions}
\end{figure}

As shown in Fig.~\ref{pearl_obs_reward_definitions}, the observation is the position of the peg relative to the noisy estimated hole position from the external sensor. The reward $r$ is defined as the negative $L_2$ distance between the peg and the actual hole, similar to \cite{metarlinsertion}, and is defined as
\begin{equation}
    r = -\| p_{\text{peg}} - p_{\text{actual hole}} \|_2,
\label{eq:reward}
\end{equation}
where $p_{\text{peg}}$ and $p_{\text{actual hole}}$ represent the positions of the peg and the actual hole in the Cartesian coordinates, respectively.

The action $\mathbf{a}$ is defined as an incremental movement of the peg up to \SI{2}{mm} in the three Cartesian coordinates
\begin{equation}
    \mathbf{a} = (\Delta x, \Delta y, \Delta z), \quad \text{with} \quad |\Delta x|, |\Delta y|, |\Delta z| \leq 2 \text{ mm},
\label{eq:action}
\end{equation}
where $\Delta x$, $\Delta y$, and $\Delta z$ denote the incremental movement in the $x$, $y$, and $z$ directions, respectively. The robot moves first in the horizontal direction and then it moves vertically, similar to \cite{metarlinsertion,visual_spatial_attention_for_concrete_peg_in_hole_assembly}, to avoid friction between the peg and the hole surface, which eliminates the need for force control.

\subsubsection{PEARL for PiH Tasks}
\label{unique_dynaimcs}
Due to the noise in the estimated hole position, the actual hole can be located at any position within a range of $\pm\delta$ around the estimated position in the horizontal plane. As shown in Fig. \ref{pearl_obs_reward_definitions}, different possible actual hole positions have different rewards. Therefore, we consider each possible actual hole position around the estimated one as a different task similar to \cite{metarlinsertion}.
However, since the observation is based on the estimated hole position, it remains the same across all tasks. Meta RL can be applied to learn how to quickly adapt to tasks with the same observation and action spaces but with different rewards and dynamics~\cite{pearl,maml,metarl_survey}. 
The dynamics, which is the next observation given the current observation and action, are unique inside the hole for each task. This is because the peg will move beneath the hole surface at a given position only if the actual hole is located at this position. Any different actual hole position, i.e., different task, will cause collisions and prevent the downwards motion of the peg.

\subsection{Context Data Modification}
\label{sec:context_data_modification}

Since the reward cannot be measured in the real-world due to the unknown actual hole position, we propose to modify the context data and use data that can be easily measured in the real world using the robot's forward kinematics and an uncalibrated camera. Additionally, using the reward in the context data for PiH assembly tasks does not provide clear information about the task as explained next.

\begin{figure}[b]

     \centering
     \begin{subfigure}[t]{0.23\textwidth}
         \centering
         \includegraphics[width=0.8\textwidth]{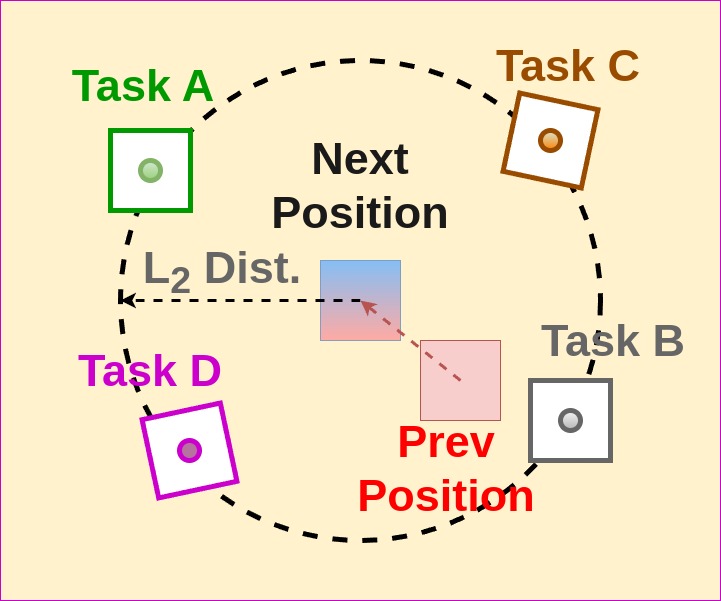}
         \caption{Tasks with the same reward are represented by the black circle.}
         \label{fig:old_context}
     \end{subfigure}
     \hfill
     \begin{subfigure}[t]{0.23\textwidth}
         \centering
         \includegraphics[width=\textwidth]{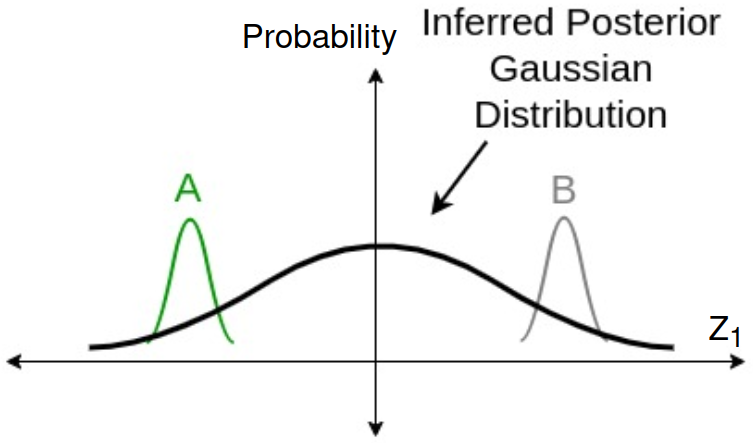}
         \caption{Posterior Gaussian distribution inferred by the context encoder.}
         \label{fig:posterior_gaussian_old_context}
     \end{subfigure}

        \caption{An example of tasks with the same reward and original context data \cite{metarlinsertion} and the corresponding posterior Gaussian distribution. Tasks with the same original context data require opposite actions resulting in a high variance posterior Gaussian distribution. }
        \label{fig:effect_of_old_context_data}
\end{figure}

\subsubsection{Effect of the Context Data on the Task Inference}
As shown in Fig. \ref{fig:old_context}, if the robot moves from the indicated previous to next positions, all tasks whose actual hole position lies on the black dashed circle will have the same reward and consequently the same context data because the tasks have the same observations. However, for example, tasks A and B require completely opposite actions by the policy. Therefore, their latent representations are far from each other, as explained in Sec. \ref{sec:PEARL}. However, since they have the same context data, the posterior Gaussian distribution produced by the context encoder tries to include both representations, resulting in a Gaussian distribution with a high variance as shown in Fig. \ref{fig:posterior_gaussian_old_context}. This degrades the ability of the agent to infer the task and consequently degrades the performance. This issue exacerbates in case of additional uncertainty in the hole orientation since tasks with the same context data may require opposite actions in three dimensions as shown by tasks C and D in Fig. \ref{fig:old_context}.

\begin{figure}[t]

     \centering
     \begin{subfigure}[t]{0.23\textwidth}
         \centering
         \includegraphics[width=0.65\textwidth]{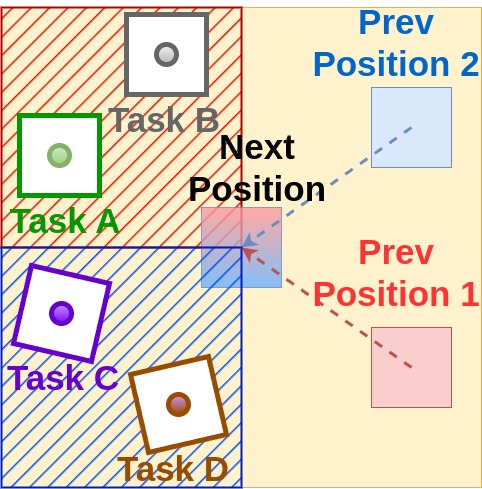}
         \caption{Space of tasks with the same $m$ is represented by the hashed square. }
         \label{fig:new_context}
     \end{subfigure}
     \hfill
     \begin{subfigure}[t]{0.23\textwidth}
         \centering
         \includegraphics[width=0.8\textwidth]{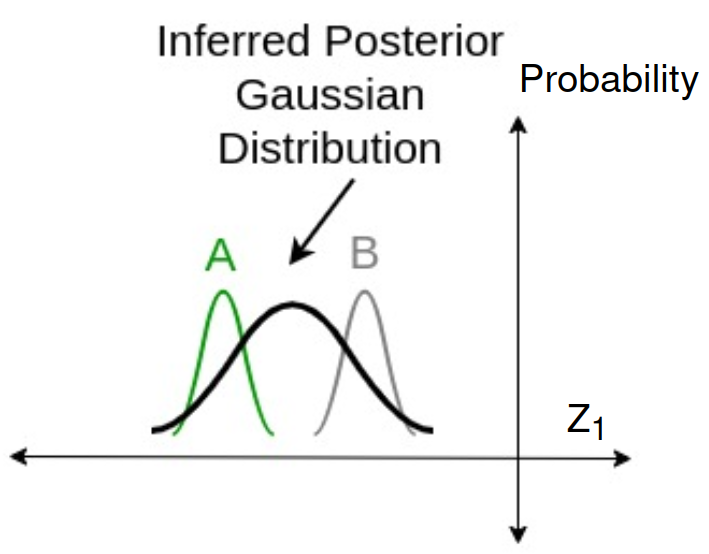}
         \caption{Posterior Gaussian distribution inferred by the context encoder. } 
         \label{fig:posterior_gaussian_new_context}
     \end{subfigure}

        \caption{An example of tasks with the same motion towards the actual hole $m$ and modified context data and the corresponding posterior Gaussian distribution.  Tasks with the same modified context data require slightly different actions resulting in small variance posterior Gaussian distribution.}
        \label{fig:effect_of_new_context_data}
\end{figure}

\subsubsection{Proposed Modified Context Data}
\label{motion_towards_goal_calculation}
We therefore propose to replace the reward in the context data with the motion towards the actual hole~$m$ based on the previous action. The motion $m$ is the summation of the distances moved by the peg towards the actual hole position in the three Cartesian coordinates, and the angle rotated around the vertical axis towards the actual hole in case of additional uncertainty in the hole orientation. $m$ is defined as follows:
\begin{equation}
    m= \pm d_x \pm d_y \pm d_z \pm \theta_z,
\label{motion_towards_the_actual_hole}
\end{equation}
where $d_x$, $d_y$, and $d_z$ are the distances moved by the peg in X, Y, and Z directions, respectively, and $\theta_z$ is the angle rotated by the peg around the vertical axis. Each distance is positive if it decreases the distance between the peg and the actual hole and vice-versa. During training in simulation, the actual hole position is known and $m$ can be accurately calculated. During execution in the real world, the distance moved in each direction is calculated from the robot forward kinematics. The sign of each distance is positive if it decreases the distance, in terms of pixels, between the detected hole and peg features in the 2D images captured before and after the motion in that direction, as shown in Fig. \ref{fig:introductory_fig}, and negative otherwise. The sign of $d_z$ is determined based on the known vertical position of the hole. Note that this procedure does not require camera calibration and it is agnostic to the position of the camera as long as the peg and the hole features can be detected before and after the motion.

\subsubsection{Effect of the Modified Context on the Task Inference}

In addition to being easily measurable, using the motion~$m$ in the context data instead of the reward results in that tasks with the same modified context data $c=(o,a,o^\prime,m)$ are closer to each other in the latent space. For example, in Fig.~\ref{fig:new_context}, tasks A and B have the same modified context data. Since they require slightly different actions, their latent representations are close to each other in the latent space resulting in a narrower Gaussian distribution by the context encoder~(see Fig.~\ref{fig:posterior_gaussian_new_context}), and the agent becomes more certain about the suitable behavior. In case of additional uncertainty in the hole orientation, tasks with the same modified context, as tasks C and D, may require opposite actions in one dimension, however, it remains more informative than the original context data with opposite action in three dimensions.

\subsection{Adaptation to Different Sensor Data}
\label{sec:adaptation_to_different_sensor}

To enhance the agent's ability to perform PiH assembly tasks where the peg occludes the hole or the camera is not available, we adapt the meta-trained agent to use data from a force/torque sensor to perform the PiH assembly task by training an alternative context encoder. 


\begin{figure}[t]

\centerline{\includegraphics[scale=0.2]{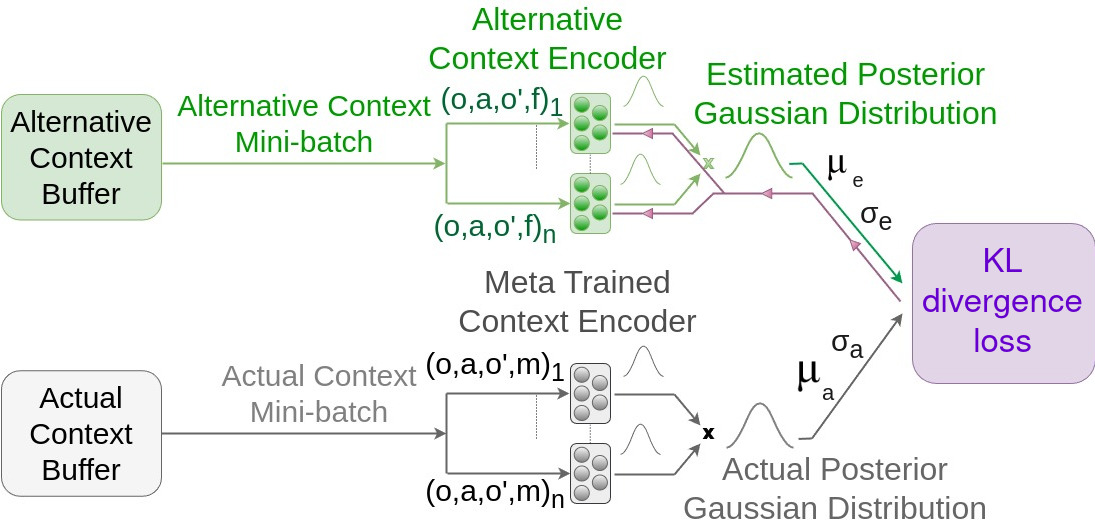}}
\caption{Alternative encoder training. Random batches of actual and alternative context data are fed to the corresponding encoder and the KL-divergence between the actual and estimated posterior distributions is calculated for each task. The KL-divergence loss is back-propagated (violet arrows) through the alternative context encoder to train and optimize it.}
\label{alternative_context_training}
\end{figure}

\subsubsection{Training Data Collection}
To collect the training data, we apply the meta-trained agent on a number of real-world tasks with different hole positions. We use latent variables sampled from the latent space of the meta training tasks to ensure safe and diverse actions by the policy. The collected actual context data $(o,a,o^\prime,m)$ and the alternative context data~$(o,a,o^\prime,f)$, where $f$ is the force/torque sensor readings, from each task are stored in separate buffers.

\subsubsection{Training Procedure}
During each iteration, random batches are independently sampled from the buffers of each task and fed to the corresponding encoder~(see Fig.~\ref{alternative_context_training}). The outputs of each batch are multiplied to produce the actual and estimated posterior Gaussian distributions for each task. The new alternative context encoder is trained to minimize the KL-divergence between the two posterior distributions of all tasks.

\begin{figure*}[t]

\begin{subfigure}{.22\textwidth}
\centering
\includegraphics[width=1.\textwidth]{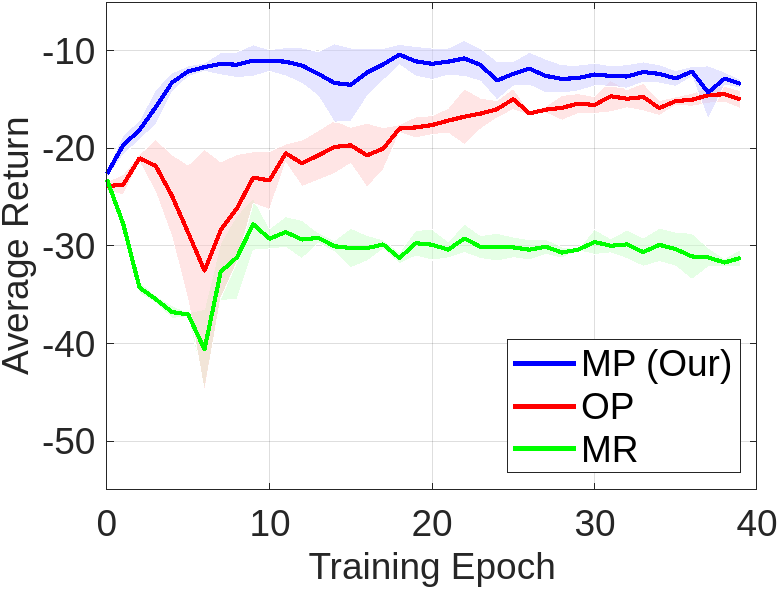}
\caption{Training tasks.} 
\label{fig: effect of context modification on training tasks}
\end{subfigure}\hspace{0.5cm}
\begin{subfigure}{.22\textwidth}
\centering
\includegraphics[width=1.\textwidth]{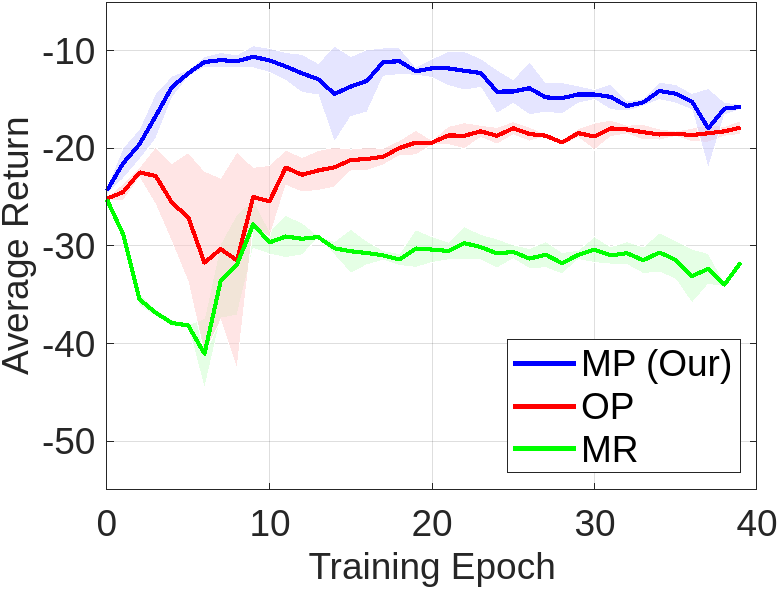}
\caption{In-distribution test tasks.}
\label{fig: effect of context modification on indistribution tasks}

\end{subfigure}\hspace{0.5cm}
\begin{subfigure}{.22\textwidth}
\centering
\includegraphics[width=1.\textwidth]{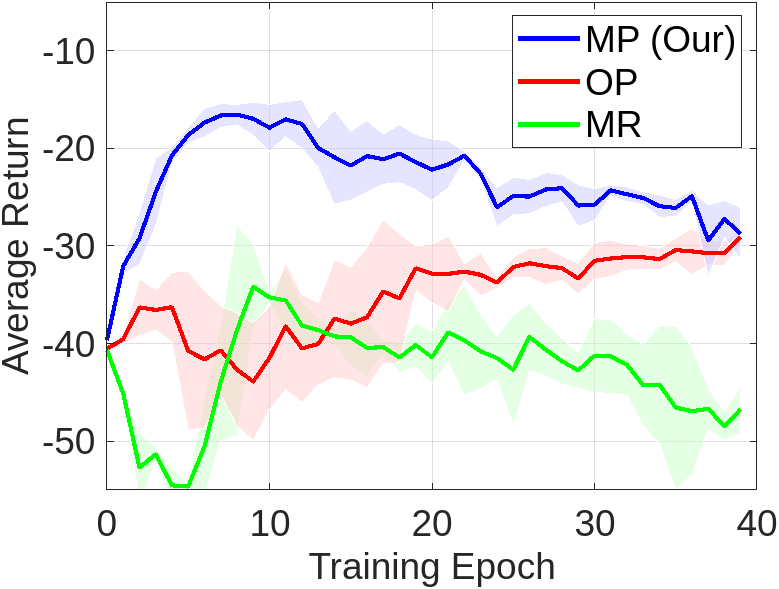}
\caption{Out-of-distribution tasks.}
\label{fig: effect of context modification on out of distribution tasks}

\end{subfigure}\hspace{0.5cm}
\begin{subfigure}{.23\textwidth}
\centering

\includegraphics[width=0.68\textwidth]{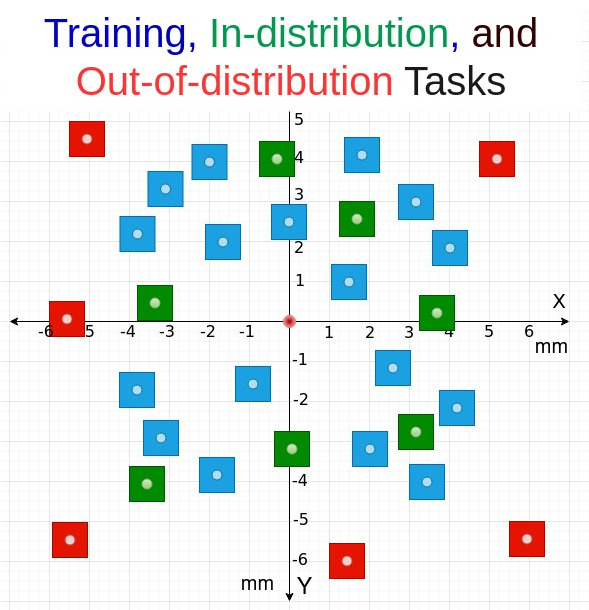}

\caption{Actual hole position of the tasks relative to the estimated position.}
\label{fig:traingvstesttasks}

\end{subfigure}
\caption{The average returns of our modified PEARL (MP), the original PEARL (OP), and the PEARL with modified reward~(MR) agents across the training epochs. As can be seen, our MP agent achieves the highest performance in all tasks compared to the other agents and requires the least number of epochs to achieve its best performance. The actual position of all tasks relative to the estimated hole in millimeters is represented in (d).
}
\label{fig: effect of context modification on meta training}

\end{figure*}

Using the posterior Gaussian distributions for the loss calculation, rather than the output of each individual sample, enables the new encoder to assign a different variance to each data sample in the batch. For example, non-informative samples with no contact or no overlapping between the peg and the hole should be assigned a high variance by the encoder so that they barely affect the posterior distribution.

\subsection{Out-of-Distribution~(OOD) Adaptation}
\label{out_of_distribution_adaptation}


 Similar to \cite{metarlinsertion}, we consider an error of $\pm$\,\SI{4}{mm} in the actual hole position. Thus, tasks with more error in the actual hole position will be considered as OOD tasks. We adapt to OOD tasks by first sampling exploration data using the normal prior distribution over the latent space and inferring the posterior distribution. This step leverages the agent's previous experience to achieve the best possible initial behavior in OOD tasks. Then, we apply an iterative process of collecting exploration data using latent variables sampled from the current posterior Gaussian distribution and optimizing the context encoder network to push the posterior Gaussian distribution towards latent representations with high values of motion towards the actual hole $m$. The context encoder is optimized to minimize the following loss:
\begin{equation}
 L_\text{OOD}=  - \alpha_1 \sum_{n=1}^N (prob(z_n) \cdot m_n ) +  \alpha_2  \frac{1}{\sigma^2}  + \alpha_3  (\mu - \text{sg}(\mu_i))^2,
\label{out_of_distribution_loss}
\end{equation}
where $N$ is the number of data sampled from the current posterior distribution, $z_n$ is the sampled latent variable at step $n$ and $m_n$ is the resulted motion towards the actual hole at the same step, $\mu$ and $\sigma^2$ are the mean and variance of the posterior Gaussian distribution produced by the context encoder during the current optimization step using the recently collected N samples, $\mu_i$ is the mean of the posterior Gaussian distribution before applying the optimization step, i.e., the posterior distribution resulting from the previous optimization step, and sg denotes to stop gradient, and $\alpha_1$, $\alpha_2$, and $\alpha_3$ are tuning weights. The first term in Eq. \ref{out_of_distribution_loss} increases the probability of latent variables with high motion towards the actual hole and vice versa. Using this term alone may severely decrease the variance, in an attempt to increase the probability, which negatively affects the exploration in the subsequent steps. Therefore, we add the second term to penalize the small variance. The third term prevents large changes in the posterior distribution between successive optimization steps to ensure safety and stability and prevent unpredictable behavior.

\begin{figure*}[t]

\begin{subfigure}{.22\textwidth}
\centering
\includegraphics[width=1\textwidth]{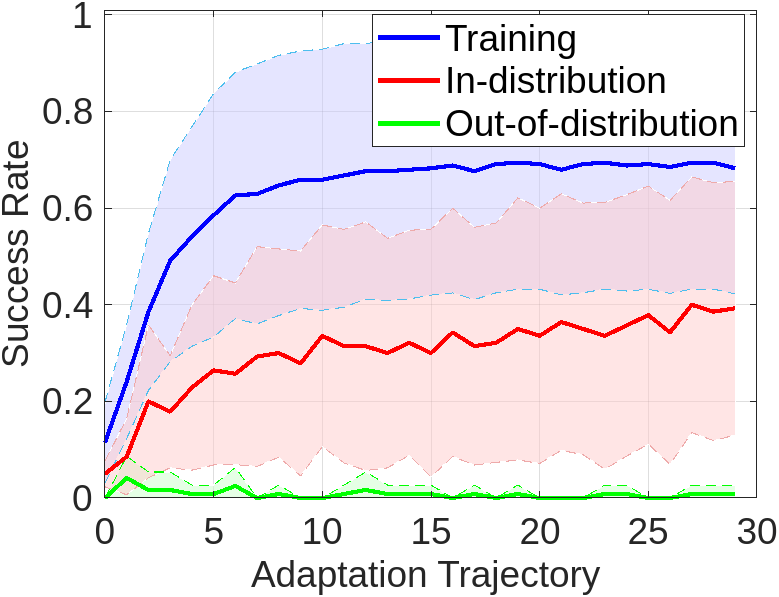}
\caption{Trajectory-based adaptation of the OP agent.}
\label{fig:succes_rate_original_context}
\end{subfigure}\hspace{0.5cm}
\begin{subfigure}{.22\textwidth}
\centering
\includegraphics[width=1\textwidth]{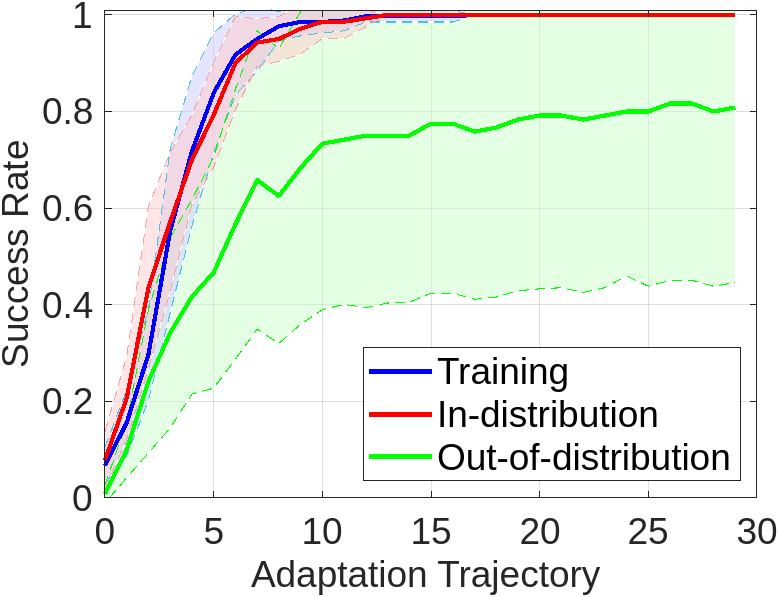}
\caption{Trajectory-based adaptation of our MP agent.}
\label{fig:Success rate of the new context agent}

\end{subfigure}\hspace{0.5cm}
\begin{subfigure}{.22\textwidth}
\centering
\includegraphics[width=1\textwidth]{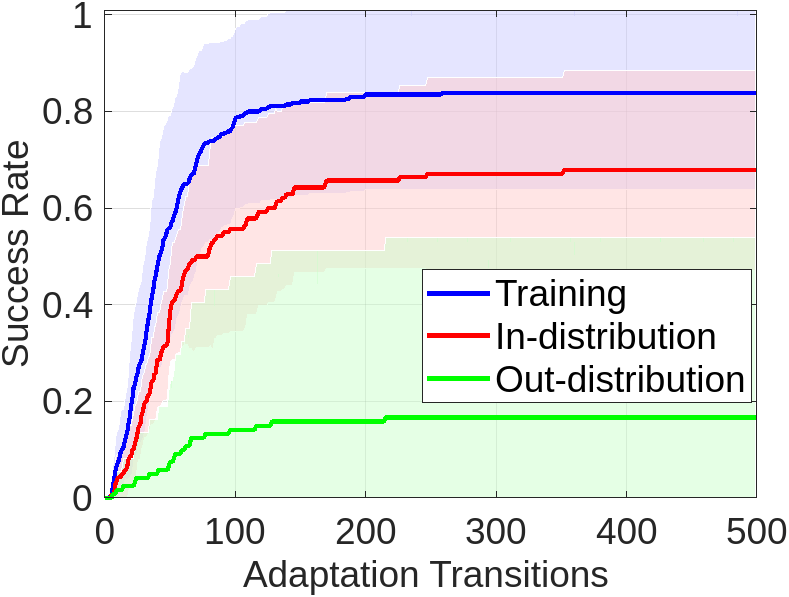}
\caption{Single transition based adaptation of the OP agent.}
\label{fig:single_transition_Adaptation_old_context}

\end{subfigure}\hspace{0.5cm}
\begin{subfigure}{.22\textwidth}
\centering
\includegraphics[width=1\textwidth]{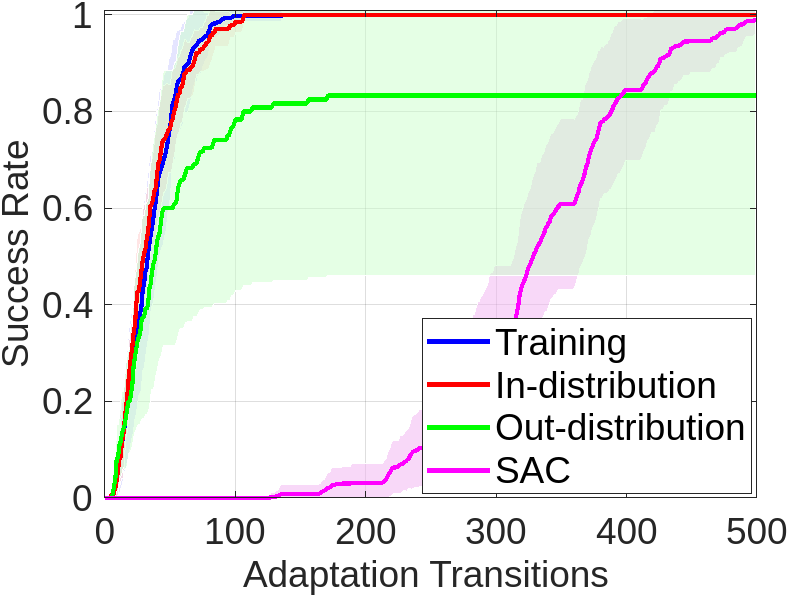}
\caption{Single transition based adaptation of our MP agent.}
\label{fig:single_transition_adaptation_new_context}

\end{subfigure}
\caption{Success rates of our modified PEARL (MP) (b,d) and the original PEARL (OP) (a,c) agents in simulation tasks. As can be seen, our MP agent outperforms the OP agent in all tasks and it is more sample-efficient than SAC.
}
\label{fig:effect_of_context_modification_on_adaptation_performance}

\end{figure*}

\section{Experimental Results}
\label{sec:results}

\par
Our experiments are designed to prove the following:
\begin{enumerate}
    \item The proposed context data modification enhances the training efficiency, adaptation performance, and robustness of PEARL in PiH assembly tasks.
    \item The proposed modifications and methods enhance the real-world adaptation performance of PEARL in PiH assembly tasks while being applicable to different peg shapes.
    \item The proposed OOD adaptation procedure enables PEARL to successfully and safely perform PiH assembly tasks with a high uncertainty in the hole position. 
    
\end{enumerate}

Our simulation experiments are carried out using \mbox{MuJoCo}~\cite{mujoco} with known actual hole positions, similar to~\cite{metarlinsertion}. Thus, the meta RL agents have access to the accurate values of the reward $r$ and the motion towards the actual hole $m$. For real-world experiments, we use the UR5e robotic arm with a built-in force/torque sensor and the D405 RealSense camera mounted on the end-effector to capture images before and after the action and we use OpenCV~\cite{opencv} to detect the hole and peg features in the captured 2D images. The starting position of the peg is 10 mm above the given estimated hole position and we assume an uncertainty of $\pm$\,\SI{4}{mm} in the actual hole positions, Similar to \cite{metarlinsertion}.

The incremental peg movement in Cartesian coordinates as output of the meta RL agent is converted into joint torque commands via MoveIt \cite{moveit} and ROS (Robot Operating System) \cite{ros}. The policy, the context encoder, and the alternative context encoder are multi-layer perceptrons (MLPs) with three hidden layers of 200 neurons each. The latent space has two dimensions for tasks with uncertainty in the hole position and five dimensions for tasks with additional uncertainty in the hole orientation.

\subsection{Effect of the Proposed Context Data Modification}

We trained two PEARL agents with identical architectures and hyper-parameters, differing only in their context data. The original PEARL (OP) \cite{metarlinsertion} uses the original context data $c=(o,a,o^{'},r)$, where $r$ is the negative $L_2$ distance reward, while our modified PEARL (MP) uses the modified context data 
$c=(o,a,o^{'},m)$, where $m$ is the motion towards the actual hole. Each agent is trained using three random seeds. The actual hole positions relative to the estimated hole position for training, in-distribution, and out-of-distribution test tasks are shown in Fig.~\ref{fig:traingvstesttasks}.

\begin{figure}[b]
     \centering
     \begin{subfigure}[t]{0.235\textwidth}
         \centering
         \includegraphics[width=1\textwidth]{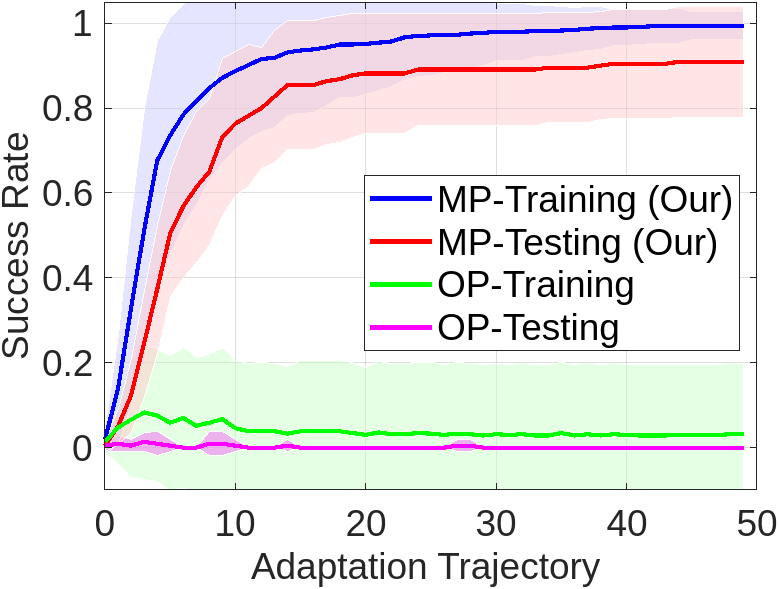}
         \caption{Trajectory adaptation.  }
         \label{fig:success_rate_orientation_traj}
     \end{subfigure}
     \hfill
     \begin{subfigure}[t]{0.235\textwidth}
         \centering
         \includegraphics[width=1\textwidth]{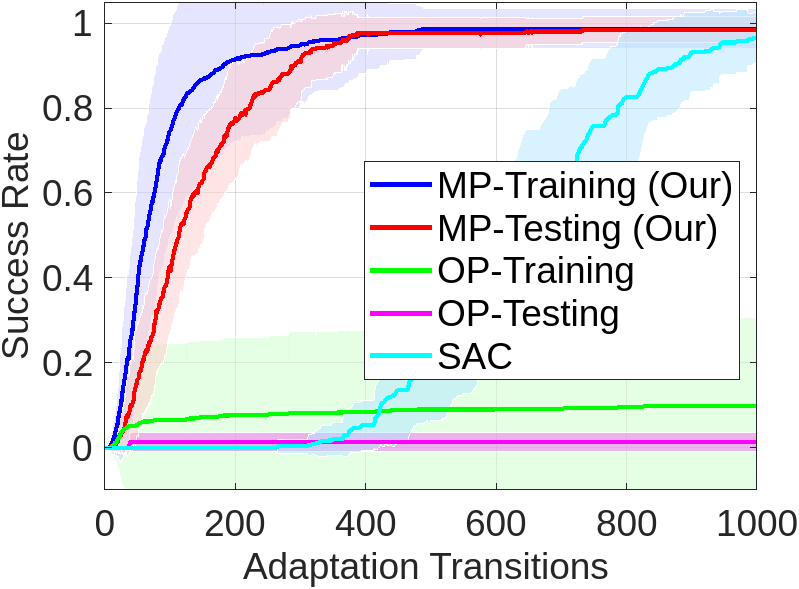}
         \caption{Single transition adaptation.   }
         \label{fig:success_rate_orientation_single_transition}
     \end{subfigure}
          
    \caption{Success rates of our modified PEARL (MP) agent and the original PEARL (OP) agent in tasks with uncertainty in the hole position and orientation. Our MP agent significantly outperforms the OP agent in all tasks, and is more sample-efficient than SAC.}
    \label{fig:adaptation_performance_orientation}
\end{figure}

\subsubsection{Training Efficiency}

We tested all agents after each epoch in all tasks for 10 trials, each trial consists of 10~successive adaptation trajectories, and calculated the average return. Return is the summation of rewards across the trajectory. The average return of each agent's random seeds across the training epochs is represented by solid lines in Fig.~\ref{fig: effect of context modification on meta training}. The shaded regions represent the standard deviation. The results indicate that our MP agent achieves higher performance than the OP agent and requires less number of epochs to achieve its best performance. This proves that the modified context data enhances the training efficiency.

The degradation of our MP agent performance, especially in test tasks in Fig.~\ref{fig: effect of context modification on indistribution tasks} and Fig.~\ref{fig: effect of context modification on out of distribution tasks}, occurs because the agent samples more context data inside the hole, which is unique to each task~(as detailed in Sec.~\ref{unique_dynaimcs}), as its performance improves. The context encoder relies on this unique data to infer the task and gradually neglects the context data outside the hole, which is similar across multiple tasks as shown in Fig.~\ref{fig:new_context}, by assigning a high variance to it.  

We trained a third agent (MR) that uses $m$ as the reward, instead of the negative $L_2$ distance, in addition to using it in the modified context data. The results in Fig.~\ref{fig: effect of context modification on meta training} show that using $m$ as a reward degrades the agent's performance. That is why our MP agent uses $m$ in the context data only and uses the  negative $L_2$ distance as a reward to train the policy.

\subsubsection{Adaptation Efficiency}

The best instance of each agent is tested in all tasks in simulation with known actual hole positions and consequently known rewards $r$ and motion~$m$. The average success rate of each agent in each task type is represented in Fig.~\ref{fig:effect_of_context_modification_on_adaptation_performance}. We adopt both trajectory-based adaptation, which updates the posterior distribution after each trajectory consisting of 50 steps \cite{metarlinsertion}, and single transition based adaptation, which updates the posterior distribution after each step. The success rate in each task is calculated based on 20 trials. The results show that our MP agent achieves a success rate of 100\,\% in all the training and test tasks within nearly 100 steps outperforming the OP agent. Even in OOD tasks, our MP agent is still able to achieve 80\,\% success rate while the OP fails. These results prove that even if the reward is accurately known, our modified context data enhances the agent's ability to infer the task and achieves more efficient adaptation.

We additionally compare our MP agent with a single RL agent trained from scratch using SAC \cite{sac}. The agent uses the motion $m$ as a reward instead of the negative $L_2$ distance, because it is available to the agent during test time in the real world. 
The results show that our MP agent is five times more sample-efficient than SAC. This proves the advantage of using meta RL for faster adaptation to new tasks than learning from scratch.

\begin{figure}[b]
     \centering
     \begin{subfigure}[t]{0.235\textwidth}
         \centering
         \includegraphics[width=0.75\textwidth]{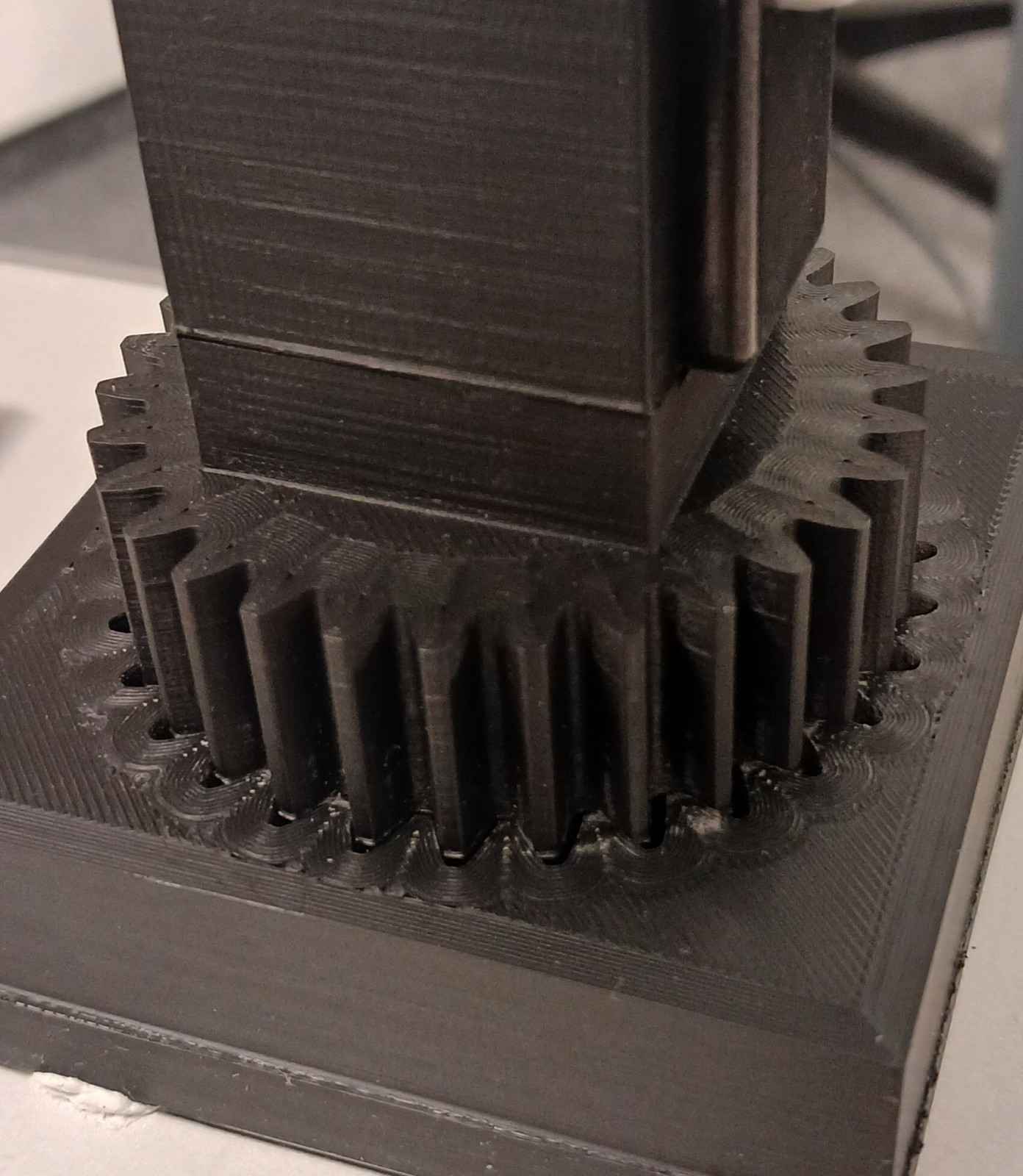}
         \caption{Gear peg/ Gear hole (GPGH).  }
         \label{fig:GPGH_configuration}
     \end{subfigure}
     \hfill
     \begin{subfigure}[t]{0.235\textwidth}
         \centering
         \includegraphics[width=0.685\textwidth]{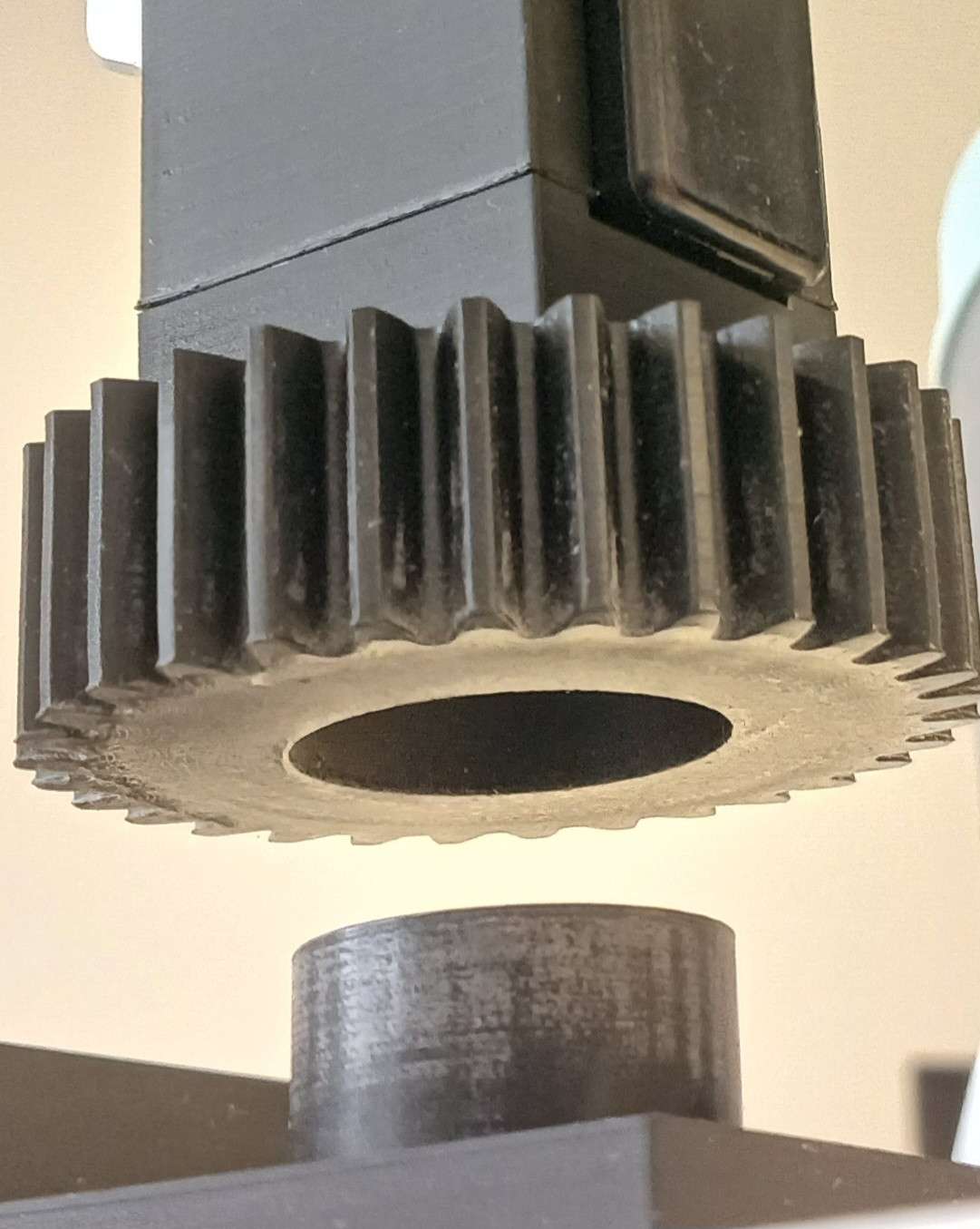}
         \caption{Gear peg/ Circular shaft (GPCS).   }
         \label{fig:GPCH configuration}
     \end{subfigure}
          
    \caption{Different peg and hole shapes used for the real-world experiments. }
    \label{fig:different_peg_shapes}
\end{figure}

\begin{figure*}[t]

\begin{subfigure}{.3\textwidth}
\centering
\includegraphics[width=0.8\textwidth]{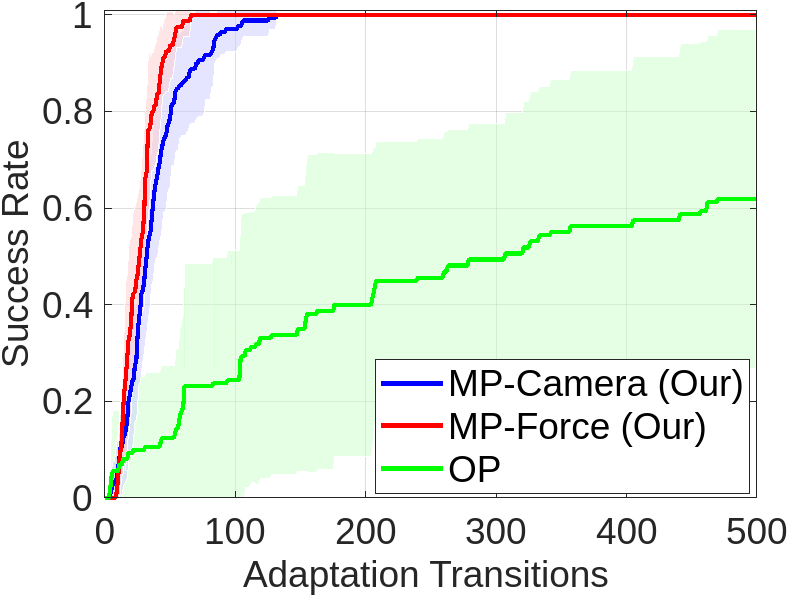}
\caption{Training tasks.} 
\label{fig: real_robot_performance_training_tasks}
\end{subfigure}\hspace{1cm}
\begin{subfigure}{.3\textwidth}
\centering
\includegraphics[width=0.8\textwidth]{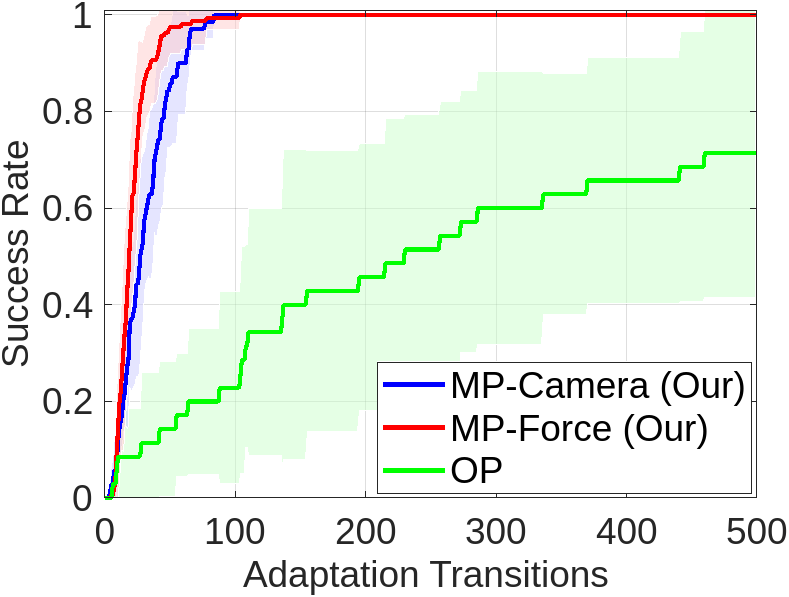}
\caption{In-distribution test tasks.}
\label{fig: real_robot_performance_indistibution_tasks}

\end{subfigure}\hspace{1cm}
\begin{subfigure}{.3\textwidth}
\centering
\includegraphics[width=0.8\textwidth]{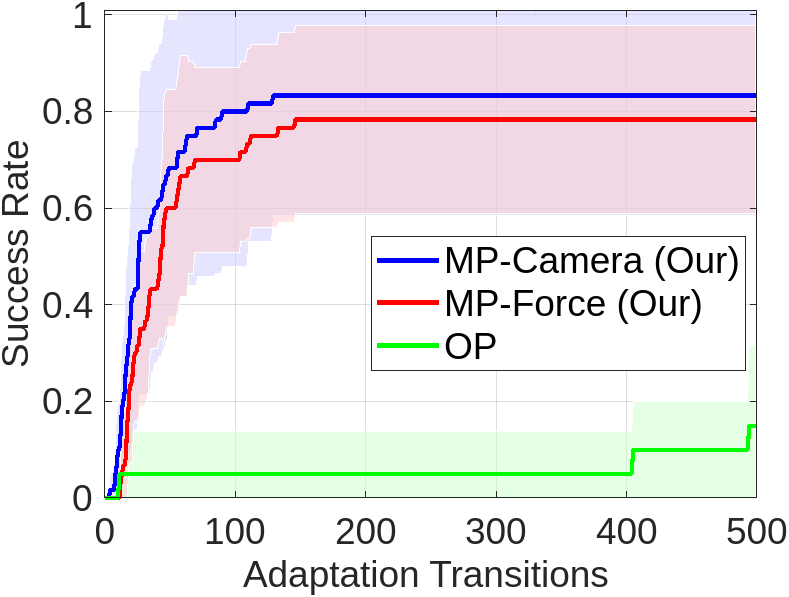}
\caption{Out-of-distribution tasks.}
\label{fig: real_robot_performance_out_distribution}

\end{subfigure}
\caption{Success rates of our modified PEARL agents using a camera (MP-Camera) and a force/torque sensor (MP-Force) and the original PEARL (OP) agents in the real-world CPSH PiH assembly task. As can be seen, our MP agents outperform the OP agent in all the tasks.
}
\label{fig: single_trasition_based_adaptation_performance}

\end{figure*}

\begin{figure}[b]
     \centering
     \begin{subfigure}[t]{0.235\textwidth}
         \centering
         \includegraphics[width=\textwidth]{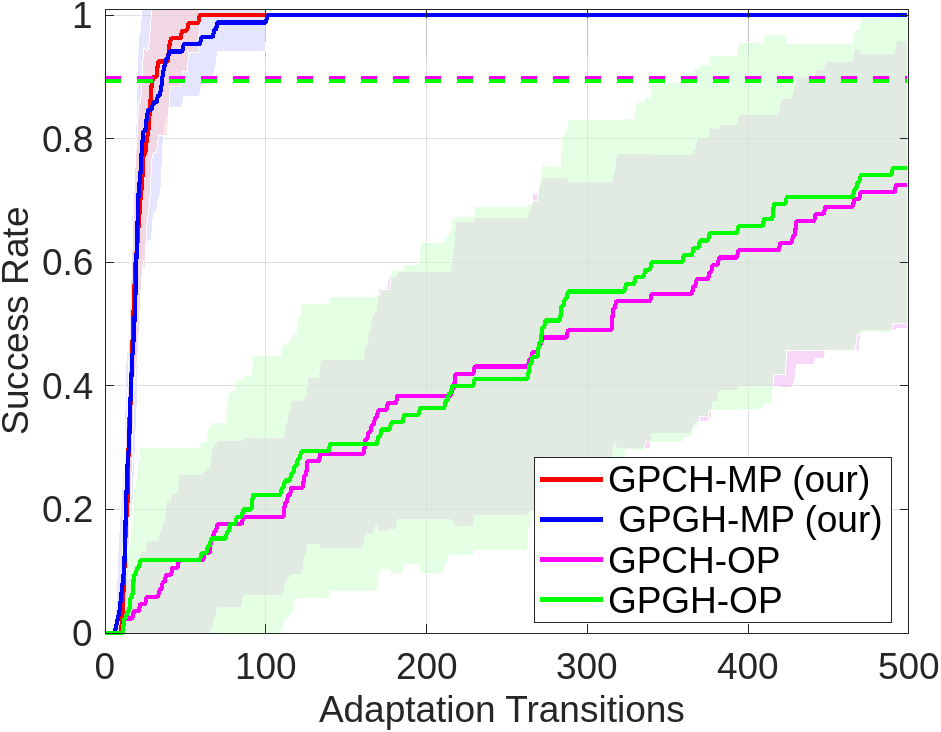}
         \caption{Training tasks  }
         \label{fig:complex_shape_training_tasks}
     \end{subfigure}
     \hfill
     \begin{subfigure}[t]{0.235\textwidth}
         \centering
         \includegraphics[width=\textwidth]{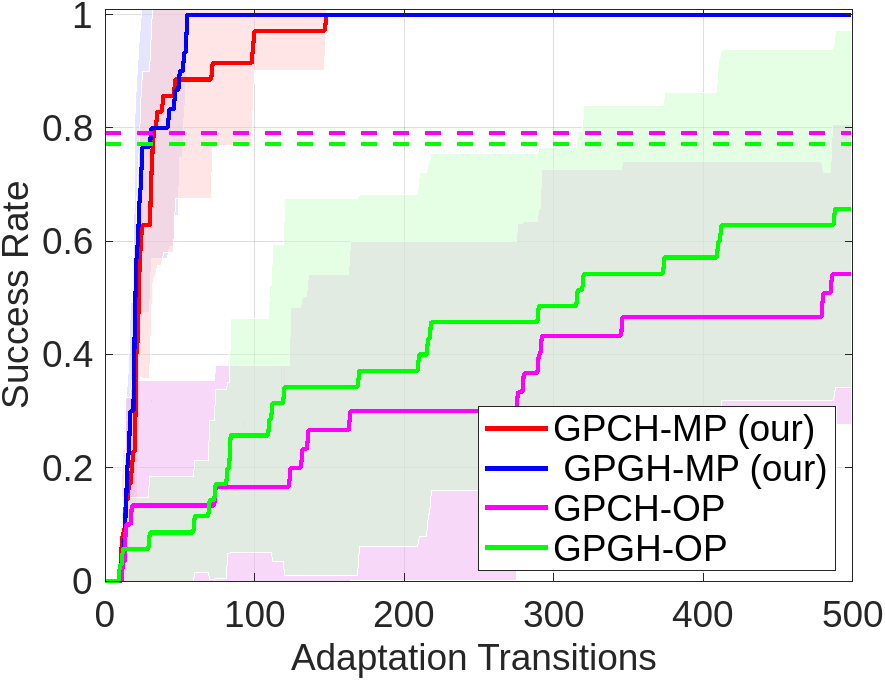}
         \caption{In-distribution test tasks.}
         \label{fig:complex_shape_indist_task}
     \end{subfigure}
          
    \caption{Success rates of our modified PEARL agents using a force/torque sensor and the original PEARL (OP) agents in real-world GPGH and GPCS PiH assembly tasks. Our MP agents outperform the OP agent in all tasks and peg shapes~(see \figref{fig:different_peg_shapes}).}
    \label{fig:real_world_performance_complex_shapes}
\end{figure}

\subsubsection{Robustness to Uncertainty in Hole Orientation}
We adapted the agents to handle additional uncertainty in the hole orientation, i.e., $\pm\,15^\circ $ around the vertical axis. To achieve this, we introduced an additional dimension representing orientation into the estimated and actual hole positions, the peg position, and the action. Furthermore, we modified the reward function and the motion towards the actual hole to account for this orientation uncertainty. The  success rates of the OP and MP agents after training for 100~epochs are represented in Fig.~\ref{fig:adaptation_performance_orientation}. The OP agent fails to learn the task achieving only 10\,\% success rate, while our MP agent achieves 100\,\% success rate in all tasks, except for 90\,\%~success rate in test tasks with the trajectory-based adaptation. Similar to our previous results, our MP agent is more sample-efficient during adaptation than the SAC agent trained from scratch. These results prove that the modified context data enhances the agent's ability to infer tasks and its robustness to more sources of uncertainty. The results show that even if the reward is known, the OP agent will not succeed, which proves the importance of our context data modification.

\begin{figure}[b]
     \centering
     \begin{subfigure}[t]{0.23\textwidth}
         \centering
         \includegraphics[width=1\textwidth]{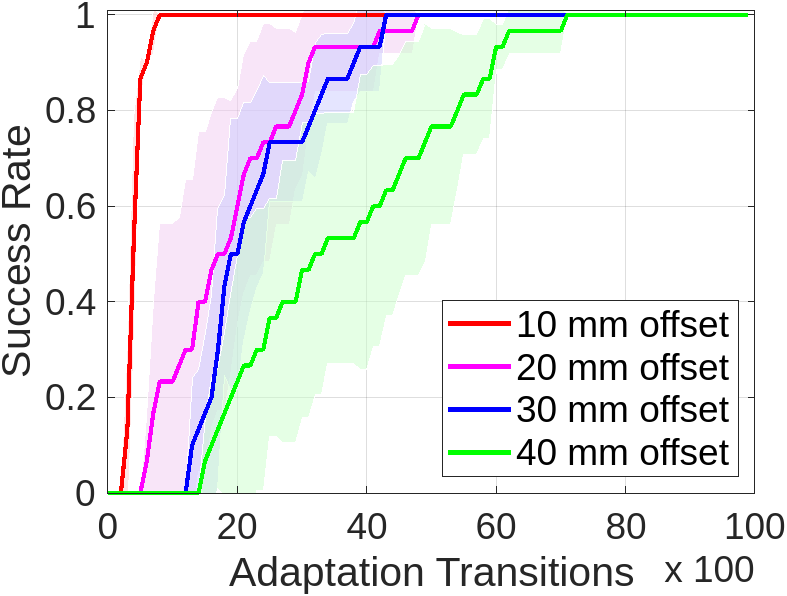}
         \caption{Success rate.  }
         \label{fig:out_of_distribution_success_rate}
     \end{subfigure}
     \hfill
     \begin{subfigure}[t]{0.23\textwidth}
         \centering
         \includegraphics[width=0.75\textwidth]{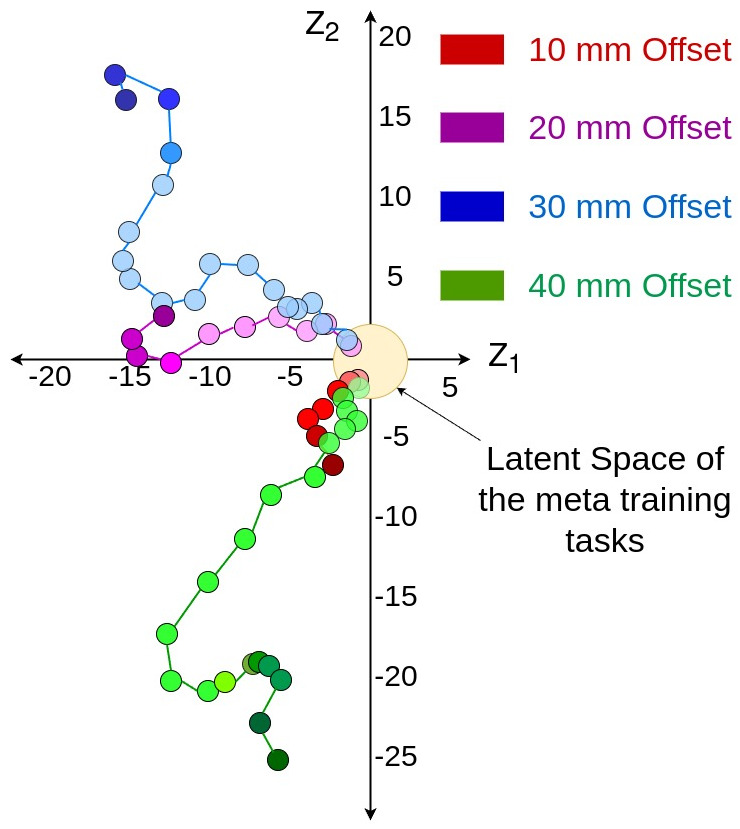}
         \caption{Exploration trajectories in the latent space.}
         \label{fig:ood_exploration_trajs}
     \end{subfigure}
          
    \caption{Success rate of our proposed adaptation method in OOD tasks (a) and examples of the resulting exploration behavior in the latent space (b). Our method consistently and gradually explores the latent space till success.}
    \label{fig:ood_results}
\end{figure}

\subsection{Real-World Adaptation Performance}

After meta training in simulation, we tested the OP, using the sparse reward trajectory-based adaptation procedure in \cite{metarlinsertion}, and our MP agents in real-world PiH with peg shapes of different complexity: Cuboid Peg in Square Hole~(CPSH) (see Fig. \ref{fig:introductory_fig}), Gear-shaped Peg in Gear-shaped Hole~(GPGH) (see Fig. \ref{fig:GPGH_configuration}), and hollow Gear Peg in Circular-shaped gear Shaft~(GPCS) (see Fig. \ref{fig:GPCH configuration}). The actual hole positions of the tasks relative to the estimated position are represented in Fig.~\ref{fig:traingvstesttasks}. 
We adapt our MP agent to use force/torque sensor data using the method proposed in Sec.~\ref{sec:adaptation_to_different_sensor} using 16,000 real-world samples for each of the three configurations, which is less than the data collected from one task per epoch during meta training, which is 32,000 samples.

 Fig.~\ref{fig: single_trasition_based_adaptation_performance} shows the average success rate of our MP agent using the end-effector camera, our MP agent using force/torque sensor data, and the OP agent in CPSH tasks. Fig. \ref{fig:real_world_performance_complex_shapes} shows the average success rate of our MP agent using force/torque sensor and the OP agent in GPGH and GPCS tasks, the dashed vertical lines show the success rate of the OP agents after 1,000 adaptation steps.  The results show that our MP agents achieves 100\,\% success rate in all the training and in-distribution test tasks within 150 adaptation steps even for complex shapes such as GPGH tasks. The OP agent achieves a maximum success rate of 80\,\% after 500 adaptation steps and 90\,\% after 1,000 steps. This proves that the OP is capable of performing the task, however, it is sample in-efficient. Additionally, the OP agents show inconsistent adaptation behavior as represented by the high standard deviations. The performance of our MP agent in the real world is nearly similar to its performance in simulation, shown in Fig.~\ref{fig:effect_of_context_modification_on_adaptation_performance}, which proves that our MP agent does not suffer from the sim-to-real gap. Compared to the results in~\cite{metarlinsertion}, which requires 20~adaptation trajectories each consisting of 50~steps to adapt to tasks with unknown hole position, our MP agents are seven times more sample-efficient during real-world adaptation.

\subsection{Out-of-Distribution Adaptation}

We applied the adaptation procedure proposed in \mbox{Sec.~\ref{out_of_distribution_adaptation}} to OOD tasks whose actual hole offsets are in the range of 10, 20, 30, and \SI{40}{mm} from the estimated hole position. The number of steps sampled per each posterior distribution $N$ is 100 transitions. The success rate of the agent is represented in Fig.~\ref{fig:out_of_distribution_success_rate}. The results show that the agent is able to achieve 100\,\% success in tasks with actual hole offsets which are 10 times greater than those used during meta training in 7,000 adaptation steps. Examples of the posterior distribution evolution  in different tasks are shown in Fig.~\ref{fig:ood_exploration_trajs}, which show how the proposed method gradually and safely explores the latent space till success.

\section{Conclusion}
\label{sec:conclusion}

In this paper, we presented novel methods to advance context-based meta RL capabilities for robust, efficient, and adaptable real-world peg-in-hole~(PiH) assembly tasks. Our modifications enhance the efficiency and applicability of the PEARL~meta RL~algorithm to perform PiH assembly tasks with unknown hole position. To this end, we proposed modified context data that can be easily measured in the real world. Additionally, we introduced a method to efficiently adapt the meta RL agent to use force/torque sensor to perform PiH assembly where data from the vision sensor is not helpful due to occlusions. Finally, we showed how to use the modified context data to adapt the meta RL to out-of-distribution tasks whose parameters are different from the training tasks.

Our experimental evaluation in simulation as well as in the real world demonstrates the superior performance of our modifications compared to the original PEARL algorithm regarding the training efficiency, adaptation performance, robustness, and generalization capabilities.

\bibliographystyle{IEEEtran}
\bibliography{bibliography}

\end{document}